\documentclass[journal]{IEEEtran}
\usepackage{amsmath}
\usepackage{amssymb}
\usepackage{array}
\usepackage{bm}
\usepackage{multirow}

\usepackage{graphicx} % more modern
\usepackage{subfigure}
\usepackage{url}

\ifCLASSINFOpdf
\else
\fi
\begin{document}
\title{A Kernel Classification Framework for Metric Learning}

\author{Faqiang~Wang,\IEEEmembership{}
        Wangmeng~Zuo,~\IEEEmembership{Member,~IEEE,}
        Lei~Zhang,~\IEEEmembership{Member,~IEEE,}
        Deyu~Meng,\IEEEmembership{}
        and~David~Zhang,~\IEEEmembership{Fellow,~IEEE}
        % <-this % stops a space
%\thanks{This work was supported in part by the National Natural Science Foundation of China under Grant 61271093 and Grant 61001037, the Hong Kong Scholar Program, and the Program of Ministry of Education for new century excellent talents.}
\thanks{F. Wang and W. Zuo are with the School of Computer Science and Technology, Harbin Institute of Technology, Harbin 150001, China (e-mail: tshfqw@163.com; cswmzuo@gmail.com).}% <-this % stops a space
\thanks{L. Zhang and D. Zhang are with the Biometrics Research Centre, Department of Computing, Hong Kong Polytechnic University, Hong Hom, Kowloon, Hong Kong (e-mail: cslzhang@comp.polyu.edu.hk; csdzhang@comp.polyu.edu.hk).}% <-this % stops a space
\thanks{D. Meng is with the Institute for Information and System Sciences, Faculty of Mathematics and Statistics, Xi'an Jiaotong University, Xi'an 710049, China (e-mail: dymeng@mail.xjtu.edu.cn).}}
%\thanks{Manuscript received April 19, 2005; revised December 27, 2012.}}


%\markboth{IEEE TRANSACTIONS ON NEURAL NETWORKS AND LEARNING SYSTEMS,~Vol.~XX, No.~X, September~2013}%
\markboth{}%
{Shell \MakeLowercase{\textit{et al.}}: Bare Demo of IEEEtran.cls for Journals}



\maketitle

\begin{abstract}
Learning a distance metric from the given training samples plays a crucial role in many machine learning tasks, and various models and optimization algorithms have been proposed in the past decade. In this paper, we generalize several state-of-the-art metric learning methods, such as large margin nearest neighbor (LMNN) and information theoretic metric learning (ITML), into a kernel classification framework. First, doublets and triplets are constructed from the training samples, and a family of degree-2 polynomial kernel functions are proposed for pairs of doublets or triplets. Then, a kernel classification framework is established, which can not only generalize many popular metric learning methods such as LMNN and ITML, but also suggest new metric learning methods, which can be efficiently implemented, interestingly, by using the standard support vector machine (SVM) solvers. Two novel metric learning methods, namely doublet-SVM and triplet-SVM, are then developed under the proposed framework. Experimental results show that doublet-SVM and triplet-SVM achieve competitive classification accuracies with state-of-the-art metric learning methods such as ITML and LMNN but with significantly less training time.
\end{abstract}


 \ifCLASSOPTIONpeerreview
 \begin{center} \bfseries EDICS Category: 3-BBND \end{center}
 \fi
\IEEEpeerreviewmaketitle

\section{Introduction}

\IEEEPARstart{H}{ow} to measure the distance (or similarity/dissimilarity) between two data points is a fundamental issue in unsupervised and supervised pattern recognition. The desired distance metrics can vary a lot in different applications due to their underlying data structures and distributions, as well as the specificity of the learning tasks. Learning a distance metric from the given training examples has been an active topic in the past decade [1], [2], and it plays a crucial role in improving the performance of many clustering (e.g., $k$-means) and classification (e.g., $k$-nearest neighbors) methods. Distance metric learning has been successfully adopted in many real world applications, e.g., face identification [3], face verification [4], image retrieval [5], [6], and activity recognition [7].
\par
Generally speaking, the goal of distance metric learning is to learn a distance metric from a given collection of similar/dissimilar samples by punishing the large distances between similar pairs and the small distances between dissimilar pairs. So far, numerous methods have been proposed to learn distance metrics, similarity metrics, and even nonlinear distance metrics. Among them, learning the Mahalanobis distance metrics for $k$-nearest neighbor classification has been receiving considerable research interests [3], [8]-[15]. The problem of similarity learning  has been studied as learning correlation metrics and cosine similarity metrics [16]-[20]. Several methods have been suggested for nonlinear distance metric learning [21], [22]. Extensions of metric learning have also been investigated for semi-supervised learning [5], [23], [24], multiple instance learning [25], and multi-task learning [26], [27], etc.
\par
Despite that many metric learning approaches have been proposed, there are still some issues to be further studied. First, since metric learning learns a distance metric from the given training dataset, it is interesting to investigate whether we can recast metric learning as a standard supervised learning problem. Second, most existing metric learning methods are motivated from specific convex programming or probabilistic models, and it is interesting to investigate whether we can unify them into a unified framework. Third, it is highly demanded that the unified framework can provide a good platform for developing new metric learning algorithms, which can be easily solved by standard and efficient learning tools.
\par
With the above considerations, in this paper we present a kernel classification framework for metric learning, which can unify most state-of-the-art metric learning methods, such as large margin nearest neighbor (LMNN) [8], [28], [29], information theoretic metric learning (ITML) [10], and logistic discriminative based metric learning (LDML) [3]. This framework allows us to easily develop new metric learning methods by using existing kernel classifiers such as the support vector machine (SVM) [30]. Under the proposed framework, we consequently present two novel metric learning methods, namely doublet-SVM and triplet-SVM, by modeling metric learning as an SVM problem, which can be efficiently solved by the existing SVM solvers like LibSVM [31].
\par
The remainder of the paper is organized as follows. Section II reviews the related work. Section III presents the proposed kernel classification framework for metric learning. Section IV introduces the doublet-SVM and triplet-SVM methods. Section V presents the experimental results, and Section VI concludes the paper.
\par
Throughout the paper, we denote matrices, vectors and scalars by the upper-case bold-faced letters, lower-case bold-faced letters, and lower-case letters, respectively.

\section{Related Work}
\par
As a fundamental problem in supervised and unsupervised learning, metric learning has been widely studied and various models have been developed, e.g., LMNN [8], ITML [10] and LDML [3]. Kumar \MakeLowercase{\textit{et al.}} extended LMNN for transformation invariant classification [32]. Huang \MakeLowercase{\textit{et al.}} proposed a generalized sparse metric learning method to learn low rank distance metrics [11]. Saenko \MakeLowercase{\textit{et al.}} extended ITML for visual category domain adaptation [33], while Kulis \MakeLowercase{\textit{et al.}} showed that in visual category recognition tasks, asymmetric transform would achieve better classification performance [34]. Cinbis \MakeLowercase{\textit{et al.}} adapted LDML to unsupervised metric learning for face identification with uncontrolled video data [35]. Several relaxed pairwise metric learning methods have been developed for efficient Mahalanobis metric learning [36], [37].
\par
Metric learning via dual approaches and kernel methods has also been studied. Shen \MakeLowercase{\textit{et al.}} analyzed the Lagrange dual of the exponential loss in the metric learning problem [12], and proposed an efficient dual approach for semi-definite metric learning [15], [38]. Actually, such boosting-like approaches usually represent the metric matrix $\mathbf{M}$ as a linear combination of rank-one matrices [39]. Liu and Vemuri  proposed a doubly regularized metric learning method by incorporating two regularization terms in the dual problem [40]. Shalev-Shwartz \MakeLowercase{\textit{et al.}} proposed a pseudo-metric online learning algorithm (POLA) to learn distance metric in the kernel space [41]. Besides, a number of pairwise SVM methods have been proposed to learn distance metrics or nonlinear distance functions [42].
\par
In this paper, we will see that most of the aforementioned metric learning approaches can be unified into the proposed kernel classification framework, while this unified framework can allow us develop new metric learning methods which can be efficiently implemented by off-the-shelf SVM tools.\\

\section{A Kernel Classification based Metric Learning Framework}
\par
Current metric learning models largely depend on convex or non-convex optimization techniques, some of which can be very inefficient to use in solving large-scale problems. In this section, we present a kernel classification framework which can unify many state-of-the-art metric learning methods, and make the metric learning task significantly more efficient. The connections between the proposed framework and LMNN, ITML, and LDML will also be discussed in detail.
\subsection{Doublets and Triplets}
\par
Unlike conventional supervised learning problems, metric learning usually considers a set of constraints imposed on the doublets or triplets of training samples to learn the desired distance metric. It is very interesting and useful to evaluate whether metric learning can be casted as a conventional supervised learning problem. To build a connection between the two problems, we model metric learning as a kind of supervised learning problem operating on a set of doublets or triplets, as described below.
\par
Let $\mathcal{D}=\left\{ \left( {{\mathbf{x}}_{i}},{{y}_{i}} \right)\left| i=1,2,\cdots ,n \right. \right\}$ be a training dataset, where vector ${{\mathbf{x}}_{i}}\in {\mathbb{R}^d}$ represents the $i$th training sample, and scalar $y_i$ represents the class label of $\mathbf{x}_i$. Any two samples extracted from $\mathcal{D}$ can form a doublet $\left( {{\mathbf{x}}_{i}},{{\mathbf{x}}_{j}} \right)$, and we assign a label $h$ to this doublet as follows: $h=-1$ if ${{y}_{i}}={{y}_{j}}$ and $h=1$ if ${{y}_{i}}\ne {{y}_{j}}$. For each training sample ${{\mathbf{x}}_{i}}$, we find from $\mathcal{D}$ its $m_1$ nearest similar neighbors, denoted by $\{\mathbf{x}_{i,1}^{s},\cdots ,\mathbf{x}_{i,{{m}_{1}}}^{s}\}$, and its $m_2$ nearest dissimilar neighbors, denoted by $\{\mathbf{x}_{i,1}^{d},\cdots ,\mathbf{x}_{i,{{m}_{2}}}^{d}\}$, and then construct $\left( {{m}_{1}}+{{m}_{2}} \right)$ doublets $\{({{\mathbf{x}}_{i}},\mathbf{x}_{i,1}^{s}),\cdots ,({{\mathbf{x}}_{i}},\mathbf{x}_{i,{{m}_{1}}}^{s}),({{\mathbf{x}}_{i}},\mathbf{x}_{i,1}^{d}),\cdots ,({{\mathbf{x}}_{i}},\mathbf{x}_{i,{{m}_{2}}}^{d})\}$. By combining all such doublets constructed from all training samples, we build a doublet set, denoted by $\{{{\mathbf{z}}_{1}},\cdots ,{{\mathbf{z}}_{{{N}_{d}}}}\}$, where ${{\mathbf{z}}_{l}}=({{\mathbf{x}}_{l,1}},{{\mathbf{x}}_{l,2}})$, $l=1,2,\cdots ,{{N}_{d}}$. The label of doublet $\mathbf{z}_l$ is denoted by $h_l$. Note that doublet based constraints are used in ITML [10] and LDML [3], but the details of the construction of doublets are not given.
\par
We call $\left( {{\mathbf{x}}_{i}},{{\mathbf{x}}_{j}},{{\mathbf{x}}_{k}} \right)$ a triplet if three samples ${{\mathbf{x}}_{i}}$, ${{\mathbf{x}}_{j}}$ and ${{\mathbf{x}}_{k}}$ are from $\mathcal{D}$ and their class labels satisfy ${{y}_{i}}={{y}_{j}}\ne {{y}_{k}}$. We adopt the following strategy to construct a triplet set. For each training sample ${{\mathbf{x}}_{i}}$, we find its $m_1$ nearest neighbors $\{\mathbf{x}_{i,1}^{s},\cdots ,\mathbf{x}_{i,{{m}_{1}}}^{s}\}$ which have the same class label as ${{\mathbf{x}}_{i}}$, and $m_2$ nearest neighbors $\{\mathbf{x}_{i,1}^{d},\cdots ,\mathbf{x}_{i,{{m}_{2}}}^{d}\}$ which have different class labels from ${{\mathbf{x}}_{i}}$. We can thus construct ${{m}_{1}}{{m}_{2}}$ triplets $\{({{\mathbf{x}}_{i}},\mathbf{x}_{i,j}^{s},\mathbf{x}_{i,k}^{d})|j=1,\cdots ,{{m}_{1}};\ k=1,\cdots ,{{m}_{2}}\}$ for each sample ${{\mathbf{x}}_{i}}$. By combining all the triplets together, we form a triplet set $\{{{\mathbf{t}}_{1}},\cdots ,{{\mathbf{t}}_{{{N}_{t}}}}\}$, where ${{\mathbf{t}}_{l}}=({{\mathbf{x}}_{l,1}},{{\mathbf{x}}_{l,2}},{{\mathbf{x}}_{l,3}})$, $l=1,2,\cdots ,{{N}_{t}}$. Note that for the convenience of expression, here we remove the super-script ``$s$'' and ``$d$'' from ${{\mathbf{x}}_{l,2}}$ and ${{\mathbf{x}}_{l,3}}$, respectively. A similar way to construct the triplets was used in LMNN [8] based on the $k$-nearest neighbors of each sample.

\subsection{A Family of Degree-2 Polynomial Kernels}
\par
We then introduce a family of degree-2 polynomial kernel functions which can operate on pairs of the doublets or triplets defined above. With the introduced degree-2 polynomial kernels, distance metric learning can be readily formulated as a kernel classification problem.
\par
Given two samples ${{\mathbf{x}}_{i}}$ and ${{\mathbf{x}}_{j}}$, we define the following function:
\begin{equation}
\label{eqn_example}
{{K}_{p}}({{\mathbf{x}}_{i}},{{\mathbf{x}}_{j}})=\operatorname{tr}({{\mathbf{x}}_{i}}\mathbf{x}_{i}^{T}{{\mathbf{x}}_{j}}\mathbf{x}_{j}^{T})
\end{equation}
where $\operatorname{tr}\left( \bullet  \right)$ represents the trace operator of a matrix. One can easily see that ${{K}_{p}}({{\mathbf{x}}_{i}},{{\mathbf{x}}_{j}})={{(\mathbf{x}_{i}^{T}{{\mathbf{x}}_{j}})}^{2}}$ is a degree-2 polynomial kernel, and ${{K}_{p}}({{\mathbf{x}}_{i}},{{\mathbf{x}}_{j}})$ satisfies the Mercer's condition [43].
\par
The kernel function defined in (1) can be extended to a pair of doublets or triplets. Given two doublets ${{\mathbf{z}}_{i}}=({{\mathbf{x}}_{i,1}},{{\mathbf{x}}_{i,2}})$ and ${{\mathbf{z}}_{j}}=({{\mathbf{x}}_{j,1}},{{\mathbf{x}}_{j,2}})$, we define the corresponding degree-2 polynomial kernel as
\setlength{\arraycolsep}{0.0em}
\begin{equation}
\begin{aligned}
 {{K}_{p}}({{\mathbf{z}}_{i}},{{\mathbf{z}}_{j}})&=\operatorname{tr}\left(
\begin{array}{c}
({{\mathbf{x}}_{i,1}}-{{\mathbf{x}}_{i,2}}){{({{\mathbf{x}}_{i,1}}-{{\mathbf{x}}_{i,2}})}^{T}}\\
({{\mathbf{x}}_{j,1}}-{{\mathbf{x}}_{j,2}}){{({{\mathbf{x}}_{j,1}}-{{\mathbf{x}}_{j,2}})}^{T}}
\end{array}%
\right) \\
& ={{\left[ {{({{\mathbf{x}}_{i,1}}-{{\mathbf{x}}_{i,2}})}^{T}}({{\mathbf{x}}_{j,1}}-{{\mathbf{x}}_{j,2}}) \right]}^{2}}
\end{aligned}.
\end{equation}
\setlength{\arraycolsep}{5pt}
\par
The kernel function in (2) defines an inner product of two doublets. With this kernel function, we can learn a decision function to tell whether the two samples of a doublet have the same class label. In Section III-C we will show the connection between metric learning and kernel decision function learning.
\par
Given two triplets ${{\mathbf{t}}_{i}}=({{\mathbf{x}}_{i,1}},{{\mathbf{x}}_{i,2}},{{\mathbf{x}}_{i,3}})$ and ${{\mathbf{t}}_{j}}=({{\mathbf{x}}_{j,1}},{{\mathbf{x}}_{j,2}},{{\mathbf{x}}_{j,3}})$, we define the corresponding degree-2 polynomial kernel as
\begin{equation}
\label{eqn_example}
{{K}_{p}}({{\mathbf{t}}_{i}},{{\mathbf{t}}_{j}})=\operatorname{tr}\left( \mathbf{T}_{i}^{{}}\mathbf{T}_{j}^{{}} \right)
\end{equation}
where
%${{\mathbf{T}}_{i}}=\left( {{\mathbf{x}}_{i,1}}-{{\mathbf{x}}_{i,3}} \right){{\left( {{\mathbf{x}}_{i,1}}-{{\mathbf{x}}_{i,3}} \right)}^{T}}-\left( {{\mathbf{x}}_{i,1}}-{{\mathbf{x}}_{i,2}} \right){{\left( {{\mathbf{x}}_{i,1}}-{{\mathbf{x}}_{i,2}} \right)}^{T}}$,${{\mathbf{T}}_{j}}=\left( {{\mathbf{x}}_{j,1}}-{{\mathbf{x}}_{j,3}} \right){{\left( {{\mathbf{x}}_{j,1}}-{{\mathbf{x}}_{j,3}} \right)}^{T}}-\left( {{\mathbf{x}}_{j,1}}-{{\mathbf{x}}_{j,2}} \right){{\left( {{\mathbf{x}}_{j,1}}-{{\mathbf{x}}_{j,2}} \right)}^{T}}$.
\[
\begin{aligned}
{{\mathbf{T}}_{i}}&=\left( {{\mathbf{x}}_{i,1}}-{{\mathbf{x}}_{i,3}} \right){{\left( {{\mathbf{x}}_{i,1}}-{{\mathbf{x}}_{i,3}} \right)}^{T}}\\
&-\left( {{\mathbf{x}}_{i,1}}-{{\mathbf{x}}_{i,2}} \right){{\left( {{\mathbf{x}}_{i,1}}-{{\mathbf{x}}_{i,2}} \right)}^{T}},
\end{aligned}
\]
\[
\begin{aligned}
{{\mathbf{T}}_{j}}&=\left( {{\mathbf{x}}_{j,1}}-{{\mathbf{x}}_{j,3}} \right){{\left( {{\mathbf{x}}_{j,1}}-{{\mathbf{x}}_{j,3}} \right)}^{T}}\\
&-\left( {{\mathbf{x}}_{j,1}}-{{\mathbf{x}}_{j,2}} \right){{\left( {{\mathbf{x}}_{j,1}}-{{\mathbf{x}}_{j,2}} \right)}^{T}}.
\end{aligned}
\]
\par
The kernel function in (3) defines an inner product of two triplets. With this kernel, we can learn a decision function based on the inequality constraints imposed on the triplets. In Section III-C we will also show how to deduce the Mahalanobis metric from the decision function.
\subsection{Metric Learning via Kernel Methods}
\par
With the degree-2 polynomial kernels defined in Section III-B, the task of metric learning can be easily solved by kernel methods. More specifically, we can use any kernel classification method to learn a kernel classifier with one of the following two forms
\begin{equation}
\begin{aligned}
{{g}_{d}\left(\mathbf{z}\right)}=\operatorname{sgn} \left( \sum\limits_{l}{{{h}_{l}}{{\alpha }_{l}}{{K}_{p}}\left( {{\mathbf{z}}_{l}},\mathbf{z} \right)}+b \right)
\end{aligned}
\end{equation}
\begin{equation}
\begin{aligned}
{{g}_{t}\left(\mathbf{t}\right)}=\operatorname{sgn} \left( \sum\limits_{l}{{{\alpha }_{l}}{{K}_{p}}\left( {{\mathbf{t}}_{l}},\mathbf{t} \right)} \right)
\end{aligned}
\end{equation}
where $\mathbf{z}_l$, $l=1,2,\cdots ,{N}$, is the doublet constructed from the training dataset, $h_l$ is the label of $\mathbf{z}_l$, $\mathbf{t}_l$, $l=1,2,\cdots,{N}$, is the triplet constructed from the training dataset, $\mathbf{z}=\left(\mathbf{x}_{\left(i\right)},\mathbf{x}_{\left(j\right)}\right)$ is the test doublet, $\mathbf{t}$ is the test triplet, $\alpha_l$ is the weight, and $b$ is the bias.
\par
For doublet, we have
\begin{equation}
\begin{aligned}
  & \sum\limits_{l}{{{h}_{l}}{{\alpha }_{l}}\operatorname{tr}\left(
\begin{array}{c}
{({{\mathbf{x}}_{l,1}}-{{\mathbf{x}}_{l,2}}){{({{\mathbf{x}}_{l,1}}-{{%
\mathbf{x}}_{l,2}})}^{T}}} \\
{({{\mathbf{x}}_{(i)}}-{{\mathbf{x}}_{(j)}}){{({{%
\mathbf{x}}_{(i)}}-{{\mathbf{x}}_{(j)}})}}^{T}}%
\end{array}%
\right)}+b \\
 & ={{({{\mathbf{x}}_{(i)}}-{{\mathbf{x}}_{(j)}})}^{T}}\mathbf{M}({{\mathbf{x}}_{(i)}}-{{\mathbf{x}}_{(j)}})+b
\end{aligned}
\end{equation}
where
\begin{equation}
\begin{aligned}
\mathbf{M}=\sum\limits_{l}{{{h}_{l}}{{\alpha }_{l}}({{\mathbf{x}}_{l,1}}-{{\mathbf{x}}_{l,2}}){{({{\mathbf{x}}_{l,1}}-{{\mathbf{x}}_{l,2}})}^{T}}}
\end{aligned}
\end{equation}
is the matrix $\mathbf{M}$ of the Mahalanobis distance metric. Thus, the kernel decision function $g_d\left(\mathbf{z}\right)$ can be used to determine whether $\mathbf{x}_{\left(i\right)}$ and $\mathbf{x}_{\left(j\right)}$ are similar or dissimilar to each other.
\par
For triplet, the matrix $\mathbf{M}$ can be derived as follows.
\newtheorem{theorem}{Theorem}
\begin{theorem}
For the decision function defined in (5), the matrix $\mathbf{M}$ of the Mahalanobis distance metric is
\begin{equation}
\begin{aligned}
\mathbf{M}&=\sum\limits_{l}{\alpha_l\mathbf{T}_l}\\
&=\sum\limits_{l}{\alpha_l\left[
\begin{array}{c}
\left(\mathbf{x}_{l,1}-\mathbf{x}_{l,3}\right)\left(\mathbf{x}_{l,1}-\mathbf{x}_{l,3}\right)^T\\
-\left(\mathbf{x}_{l,1}-\mathbf{x}_{l,2}\right)\left(\mathbf{x}_{l,1}-\mathbf{x}_{l,2}\right)^T
\end{array}
\right]}
\end{aligned}
\end{equation}
and then $\sum\nolimits_{l}{\alpha_l K_p\left( \mathbf{t}_l,\mathbf{t} \right)}$  denotes the relative difference of the Mahalanobis distance between $\mathbf{x}_{\left(i\right)}$ and $\mathbf{x}_{\left(k\right)}$ and the Mahalanobis distance between $\mathbf{x}_{\left(i\right)}$ and $\mathbf{x}_{\left(j\right)}$.
\end{theorem}
\begin{IEEEproof}
Let $\mathbf{T}_l=\left( \mathbf{x}_{l,1}-\mathbf{x}_{l,3} \right)\left( \mathbf{x}_{l,1}-\mathbf{x}_{l,3} \right)^T-\left( \mathbf{x}_{l,1}-\mathbf{x}_{l,2} \right)\left( \mathbf{x}_{l,1}-\mathbf{x}_{l,2} \right)^T$. Based on the definition of $K_p\left( \mathbf{t}_l,\mathbf{t} \right)$, we have
\begin{equation}
\begin{aligned}
& \sum\limits_{l}{\alpha_l K_p\left( \mathbf{t}_l,\mathbf{t} \right)} = \sum\limits_{l}{\alpha_l\operatorname{tr}\left(\mathbf{T}_l\mathbf{T}\right)}\\
= & \sum\limits_{l}{\alpha_l\operatorname{tr}\left(\mathbf{T}_l\left(
\begin{array}{c}
\left(\mathbf{x}_{\left(i\right)}-\mathbf{x}_{\left(k\right)}\right)\left(\mathbf{x}_{\left(i\right)}-\mathbf{x}_{\left(k\right)}\right)^T\\
-\left(\mathbf{x}_{\left(i\right)}-\mathbf{x}_{\left(j\right)}\right)\left(\mathbf{x}_{\left(i\right)}-\mathbf{x}_{\left(j\right)}\right)^T
\end{array}
\right)^T\right)}\\
=&\sum\limits_{l}{\alpha_l\operatorname{tr}\left(\mathbf{T}_l\left(\left(\mathbf{x}_{\left(i\right)}-\mathbf{x}_{\left(k\right)}\right)\left(\mathbf{x}_{\left(i\right)}-\mathbf{x}_{\left(k\right)}\right)^T\right)^T\right)}\\
& - \sum\limits_{l}{\alpha_l\operatorname{tr}\left(\mathbf{T}_l\left(\left(\mathbf{x}_{\left(i\right)}-\mathbf{x}_{\left(j\right)}\right)\left(\mathbf{x}_{\left(i\right)}-\mathbf{x}_{\left(j\right)}\right)^T\right)^T\right)}\\
=&\left(\mathbf{x}_{\left(i\right)}-\mathbf{x}_{\left(k\right)}\right)^T\left(\sum\limits_{l}{\alpha_l\mathbf{T}_l}\right)\left(\mathbf{x}_{\left(i\right)}-\mathbf{x}_{\left(k\right)}\right)\\
& - \left(\mathbf{x}_{\left(i\right)}-\mathbf{x}_{\left(j\right)}\right)^T\left(\sum\limits_{l}{\alpha_l\mathbf{T}_l}\right)\left(\mathbf{x}_{\left(i\right)}-\mathbf{x}_{\left(j\right)}\right)\\
=&\left(\mathbf{x}_{\left(i\right)}-\mathbf{x}_{\left(k\right)}\right)^T\mathbf{M}\left(\mathbf{x}_{\left(i\right)}-\mathbf{x}_{\left(k\right)}\right)\\
& - \left(\mathbf{x}_{\left(i\right)}-\mathbf{x}_{\left(j\right)}\right)^T\mathbf{M}\left(\mathbf{x}_{\left(i\right)}-\mathbf{x}_{\left(j\right)}\right)
\end{aligned}.
\end{equation}
\par
By setting $\mathbf{M}=\sum\nolimits_{l}{\alpha_l\mathbf{T}_l}$ as the matrix $\mathbf{M}$ in the Mahalanobis distance metric, we can see that $\sum\nolimits_{l}{\alpha_l K_p\left(\mathbf{t}_l,\mathbf{t}\right)}$  is the difference of the distance between $\mathbf{x}_{\left(i\right)}$ and $\mathbf{x}_{\left(k\right)}$ and that between $\mathbf{x}_{\left(i\right)}$ and $\mathbf{x}_{\left(j\right)}$.
\end{IEEEproof}
\par
Clearly, equations (4) $\thicksim$ (9) provide us a new perspective to view and understand the distance metric matrix $\mathbf{M}$ under a kernel classification framework. Meanwhile, this perspective provides us new approaches for learning distance metric, which can be much easier and more efficient than the previous metric learning approaches. In the following, we introduce two kernel classification methods for metric learning: regularized kernel SVM and kernel logistic regression. Note that by modifying the construction of doublet or triplet set, using different kernel classifier models, or adopting different optimization algorithms, other new metric learning algorithms can also be developed under the proposed framework.\\
\subsubsection{Kernel SVM-like Model}
Given the doublet or triplet training set, an SVM-like model can be proposed to learn the distance metric:
\begin{equation}
\begin{aligned}
  & \underset{\mathbf{M},b,\boldsymbol{\xi }}{\mathop{\min }}\,\text{ }r\left( \mathbf{M} \right)+\rho \left( \boldsymbol{\xi } \right) \\
 & \text{s}\text{.t}\text{.}\quad f_{l}^{(d)}\left( {{({{\mathbf{x}}_{l,1}}-{{\mathbf{x}}_{l,2}})}^{T}}\mathbf{M}({{\mathbf{x}}_{l,1}}-{{\mathbf{x}}_{l,2}}),b,{{\xi }_{l}} \right)\ge 0\text{  }(\text{doublet set}) \\
 & \quad \quad \text{or    }f_{l}^{(t)}\left(
 \begin{array}{c}
 {{({{\mathbf{x}}_{l,1}}-{{\mathbf{x}}_{l,3}})}^{T}}\mathbf{M}({{\mathbf{x}}_{l,1}}-{{\mathbf{x}}_{l,3}})\\
 -{{({{\mathbf{x}}_{l,1}}-{{\mathbf{x}}_{l,2}})}^{T}}\mathbf{M}({{\mathbf{x}}_{l,1}}-{{\mathbf{x}}_{l,2}})
 \end{array}
 ,{{\xi }_{l}} \right)\ge 0\text{  }(\text{triplet set}) \\
 & \quad \quad {{\xi }_{l}}\ge 0 \\
\end{aligned}
\end{equation}
where $r\left( \mathbf{M} \right)$ is the regularization term, $\rho \left( \boldsymbol{\xi } \right)$ is the margin loss term, the constraint $f_{l}^{(d)}$ can be any linear function of ${{({{\mathbf{x}}_{l,1}}-{{\mathbf{x}}_{l,2}})}^{T}}\mathbf{M}({{\mathbf{x}}_{l,1}}-{{\mathbf{x}}_{l,2}})$, $b$, and ${{\xi }_{l}}$, and the constraint $f_{l}^{(t)}$ can be any linear function of ${{({{\mathbf{x}}_{l,1}}-{{\mathbf{x}}_{l,3}})}^{T}}\mathbf{M}({{\mathbf{x}}_{l,1}}-{{\mathbf{x}}_{l,3}})-{{({{\mathbf{x}}_{l,1}}-{{\mathbf{x}}_{l,2}})}^{T}}\mathbf{M}({{\mathbf{x}}_{l,1}}-{{\mathbf{x}}_{l,2}})$ and ${{\xi }_{l}}$. To guarantee that (10) is convex, we can simply choose convex regularizer $r\left( \mathbf{M} \right)$ and convex margin loss $\rho \left( \boldsymbol{\xi } \right)$. By plugging (7) or (8) in the model in (10), we can employ the SVM and kernel methods to learn all $\alpha_l$ to obtain the matrix $\mathbf{M}$.
\par
If we adopt the $l_2$-norm to regularize $\mathbf{M}$ and the hinge loss penalty on ${{\xi }_{l}}$, the model in (10) would become the standard SVM. SVM and its variants have been extensively studied [30], [44], [45], and various algorithms have been proposed for large-scale SVM training [46], [47]. Thus, the SVM-like modeling in (10) can allow us to learn good metrics efficiently from large-scale training data.
\subsubsection{Kernel logistic regression}
Under the kernel logistic regression model (KLR) [48], we let ${{h}_{l}}=1$ if the samples of doublet $\mathbf{z}_l$ belong to the same class and let ${{h}_{l}}=0$ if the samples of it belong to different classes. Meanwhile, suppose that the label of a doublet $\mathbf{z}_l$ is unknown, and we can calculate the probability that $\mathbf{z}_l$'s label is 1 as follows:
\begin{equation}
\begin{aligned}
%P({{p}_{l}}=1|{{\mathbf{z}}_{l}})&=\frac{1}{1+\exp \left( {{({{\mathbf{x}}_{l,1}}-{{\mathbf{x}}_{l,2}})}^{T}}\mathbf{M}({{\mathbf{x}}_{l,1}}-{{\mathbf{x}}_{l,2}})+b \right)}\\
%&=\frac{1}{1+\exp \left(\sum\limits_{i}{\alpha_i h_i K_p\left(\mathbf{z}_i,\mathbf{z}_l\right)} +b \right)}
P({{p}_{l}}=1|{{\mathbf{z}}_{l}})=\frac{1}{1+\exp \left(\sum\limits_{i}{\alpha_i K_p\left(\mathbf{z}_i,\mathbf{z}_l\right)} +b \right)}.
\end{aligned}
\end{equation}
\par
The coefficient $\boldsymbol{\alpha}$ and the bias $b$ can be obtained by maximizing the following log-likelihood function:
\begin{equation}
\begin{aligned}
(\boldsymbol{\alpha},b)=\arg \underset{\boldsymbol{\alpha},b}{\mathop{\max }}\,\left\{
\begin{array}{c}
l(\boldsymbol{\alpha},b)=\sum\limits_{l}{{{h}_{l}}\ln P({{p}_{l}}=1|{{\mathbf{z}}_{l}})}\\
  +(1-{{h}_{l}})\ln P({{p}_{l}}=0|{{\mathbf{z}}_{l}})
\end{array}
\right\}
\end{aligned}.
\end{equation}
\par
KLR is a powerful probabilistic approach for classification. By modeling metric learning as a KLR problem, we can easily use the existing KLR algorithms to learn the desired metric. Moreover, the variants and improvements of KLR, e.g., sparse KLR [49], can also be used to develop new metric learning methods.
\subsection{Connections with LMNN, ITML, and LDML}
\par
The proposed kernel classification framework provides a unified explanation of many state-of-the-art metric learning methods. In this subsection, we show that LMNN and ITML can be considered as certain SVM models, while LDML is an example of the kernel logistic regression model.
\subsubsection{LMNN}
LMNN [8] learns a distance metric that penalizes both large distances between samples with the same label and small distances between samples with different labels. LMNN is operated on a set of triplets $\left\{ \left( {{\mathbf{x}}_{i}},{{\mathbf{x}}_{j}},{{\mathbf{x}}_{k}} \right) \right\}$, where ${{\mathbf{x}}_{i}}$ has the same label as ${{\mathbf{x}}_{j}}$ but has different label from ${{\mathbf{x}}_{k}}$. The minimization of LMNN can be stated as follows:
\begin{equation}
\begin{aligned}
   \underset{\mathbf{M},{{\xi }_{ijk}}}{\mathop{\min }}\,\text{ } & \sum\limits_{i,j}{{{\left( {{\mathbf{x}}_{i}}-{{\mathbf{x}}_{j}} \right)}^{T}}\mathbf{M}\left( {{\mathbf{x}}_{i}}-{{\mathbf{x}}_{j}} \right)}+C\sum\limits_{i,j,k}{{{\xi }_{ijk}}} \\
  \text{        s}\text{.t}\text{. } &
  \begin{array}{c}
  {{\left( {{\mathbf{x}}_{i}}-{{\mathbf{x}}_{k}} \right)}^{T}}\mathbf{M}\left( {{\mathbf{x}}_{i}}-{{\mathbf{x}}_{k}} \right)\\
  -{{\left( {{\mathbf{x}}_{i}}-{{\mathbf{x}}_{j}} \right)}^{T}}\mathbf{M}\left( {{\mathbf{x}}_{i}}-{{\mathbf{x}}_{j}} \right)
  \end{array}
  \ge 1-{{\xi }_{ijk}} \\
 & \text{              }{{\xi }_{ijl}}\ge 0 \\
 & \text{              }\mathbf{M}\succcurlyeq 0 \\
\end{aligned}.
\end{equation}
\par
Since $\mathbf{M}$ is required to be positive semi-definite in LMNN, we introduce the following indicator function:
\begin{equation}
{{\iota }_{\succcurlyeq }}\left( \mathbf{M} \right)=\left\{ \begin{aligned}
  & 0,\quad \operatorname{if}\ \mathbf{M}\succcurlyeq 0 \\
 & +\infty ,\quad \text{otherwise} \\
\end{aligned} \right.
\end{equation}
and choose the following regularizer and margin loss:
\begin{equation}
\begin{aligned}
{{r}_{\operatorname{LMNN}}}\left( \mathbf{M} \right)=\sum\limits_{i,j}{{{\left( {{\mathbf{x}}_{i}}-{{\mathbf{x}}_{j}} \right)}^{T}}\mathbf{M}\left( {{\mathbf{x}}_{i}}-{{\mathbf{x}}_{j}} \right)}+{{\iota }_{\succcurlyeq }}\left( \mathbf{M} \right)
\end{aligned}
\end{equation}
\begin{equation}
\begin{aligned}
{{\rho }_{\text{LMNN}}}\left( \boldsymbol{\xi } \right)=C\sum\limits_{i,j,k}{{{\xi }_{ijk}}}.
\end{aligned}
\end{equation}
\par
Then we can define the following SVM-like model on the same triplet set:
\begin{equation}
\begin{aligned}
  & \underset{\mathbf{M},\boldsymbol{\xi }}{\mathop{\min }}\,\text{ }{{r}_{\operatorname{LMNN}}}\left( \mathbf{M} \right)+{{\rho }_{\operatorname{LMNN}}}\left( \boldsymbol{\xi } \right) \\
 & \text{s}\text{.t}\text{.}\quad %{{({{\mathbf{x}}_{i}}-{{\mathbf{x}}_{k}})}^{T}}\mathbf{M}({{\mathbf{x}}_{i}}-{{\mathbf{x}}_{k}})-{{({{\mathbf{x}}_{i}}-{{\mathbf{x}}_{j}})}^{T}}\mathbf{M}({{\mathbf{x}}_{i}}-{{\mathbf{x}}_{j}})
 \begin{array}{c}
  {{\left( {{\mathbf{x}}_{i}}-{{\mathbf{x}}_{k}} \right)}^{T}}\mathbf{M}\left( {{\mathbf{x}}_{i}}-{{\mathbf{x}}_{k}} \right)\\
  -{{\left( {{\mathbf{x}}_{i}}-{{\mathbf{x}}_{j}} \right)}^{T}}\mathbf{M}\left( {{\mathbf{x}}_{i}}-{{\mathbf{x}}_{j}} \right)
  \end{array}
 \ge 1-{{\xi }_{ijk}}\text{ } \\
 & \quad \quad {{\xi }_{ijk}}\ge 0 \\
\end{aligned}.
\end{equation}
\par
It is obvious that the SVM-like model in (17) is equivalent to the LMNN model in (13).
\subsubsection{ITML}
ITML [10] is operated on a set of doublets $\left\{ \left( {{\mathbf{x}}_{i}},{{\mathbf{x}}_{j}} \right) \right\}$ by solving the following minimization problem£º
\begin{equation}
\begin{aligned}
   \underset{\mathbf{M},\boldsymbol{\xi }}{\mathop{\min }}\,\text{ } & {{D}_{ld}}\left( \mathbf{M},{{\mathbf{M}}_{0}} \right)+\gamma \cdot {{D}_{ld}}\left( \operatorname{diag}\left( \boldsymbol{\xi } \right),\operatorname{diag}\left( {{\boldsymbol{\xi }}_{0}} \right) \right) \\
  \text{        s}\text{.t}\text{. } & {{({{\mathbf{x}}_{i}}-{{\mathbf{x}}_{j}})}^{T}}\mathbf{M}({{\mathbf{x}}_{i}}-{{\mathbf{x}}_{j}})\le {{\xi }_{u\left( i,j \right)}}\text{  }\left( i,j \right)\in \mathcal{S} \\
 & {{({{\mathbf{x}}_{i}}-{{\mathbf{x}}_{j}})}^{T}}\mathbf{M}({{\mathbf{x}}_{i}}-{{\mathbf{x}}_{j}})\ge {{\xi }_{l\left( i,j \right)}}\text{   }\left( i,j \right)\in \mathcal{D} \\
 & \mathbf{M}\succcurlyeq 0 \\
\end{aligned}
\end{equation}
where ${{\mathbf{M}}_{0}}$ is the given prior of the metric matrix, ${{\boldsymbol{\xi }}_{0}}$ is the given prior on $\boldsymbol{\xi }$, $\mathcal{S}$ is the set of doublets where ${{\mathbf{x}}_{i}}$ and ${{\mathbf{x}}_{j}}$ have the same label, $\mathcal{D}$ is the set of doublets where ${{\mathbf{x}}_{i}}$ and ${{\mathbf{x}}_{j}}$ have different labels, and ${{D}_{ld}}\left( \cdot ,\cdot  \right)$ is the LogDet divergence of two matrices defined as:
\begin{equation}
\begin{aligned}
{{D}_{ld}}\left( \mathbf{M},{{\mathbf{M}}_{0}} \right)=\operatorname{tr}\left( \mathbf{MM}_{0}^{-1} \right)-\log \det \left( \mathbf{MM}_{0}^{-1} \right)-n.
\end{aligned}
\end{equation}
\par
By introducing the following regularizer and margin loss:
\begin{equation}
\begin{aligned}
{{r}_{\operatorname{ITML}}}\left( \mathbf{M} \right)={{D}_{ld}}\left( \mathbf{M},{{\mathbf{M}}_{0}} \right)+{{\iota }_{\succcurlyeq }}\left( \mathbf{M} \right)
\end{aligned}
\end{equation}
\begin{equation}
\begin{aligned}
{{\rho }_{\text{ITML}}}\left( \boldsymbol{\xi } \right)=\gamma \cdot {{D}_{ld}}\left( \operatorname{diag}\left( \boldsymbol{\xi } \right),\operatorname{diag}\left( {{\boldsymbol{\xi }}_{0}} \right) \right)
\end{aligned}
\end{equation}
we can then define the following SVM-like model on the same doublet set:
\begin{equation}
\begin{aligned}
   \underset{\mathbf{M},\boldsymbol{\xi }}{\mathop{\min }}\,\text{ } & {{r}_{\text{ITML}}}\left( \mathbf{M} \right)+{{\rho }_{\text{ITML}}}\left( \boldsymbol{\xi } \right) \\
 \text{s}\text{.t}\text{.} \quad &
 {{({{\mathbf{x}}_{i}}-{{\mathbf{x}}_{j}})}^{T}}\mathbf{M}({{\mathbf{x}}_{i}}-{{\mathbf{x}}_{j}})\le {{\xi }_{u\left( i,j \right)}}\text{  }\left( i,j \right)\in \mathcal{S} \\
 & {{({{\mathbf{x}}_{i}}-{{\mathbf{x}}_{j}})}^{T}}\mathbf{M}({{\mathbf{x}}_{i}}-{{\mathbf{x}}_{j}})\ge {{\xi }_{l\left( i,j \right)}}\text{   }\left( i,j \right)\in \mathcal{D} \\
 & {{\xi }_{ij}}\ge 0 \\
\end{aligned}
\end{equation}
where ${{\mathbf{z}}_{ij}}=\left( {{\mathbf{x}}_{i}},{{\mathbf{x}}_{j}} \right)$. One can easily see that the SVM-like model in (22) is equivalent to the ITML model in (18).
\subsubsection{LDML}
LDML [3] is a logistic discriminant based metric learning approach based on a set of doublets. Given a doublet ${{\mathbf{z}}_{l}}=\left( {{\mathbf{x}}_{l(i)}},{{\mathbf{x}}_{l(j)}} \right)$ and its label $h_l$, LDML defines the probability that ${{y}_{l(i)}}={{y}_{l(j)}}$ as follows:
\begin{equation}
\begin{aligned}
{{p}_{l}} & =P({{y}_{l(i)}}={{y}_{l(j)}}|{{\mathbf{x}}_{l(i)}},{{\mathbf{x}}_{l(j)}},\mathbf{M},b)\\
 & =\sigma (b-{{d}_{\mathbf{M}}}({{\mathbf{x}}_{l(i)}},{{\mathbf{x}}_{l(j)}}))
\end{aligned}
\end{equation}
where $\sigma (z)$ is the sigmoid function, $b$ is the bias, and ${{d}_{\mathbf{M}}}({{\mathbf{x}}_{l(i)}},{{\mathbf{x}}_{l(j)}})={{({{\mathbf{x}}_{l(i)}}-{{\mathbf{x}}_{l(j)}})}^{T}}\mathbf{M}({{\mathbf{x}}_{l(i)}}-{{\mathbf{x}}_{l(j)}})$. With the $p_l$ defined in (23), LDML learns $\mathbf{M}$ and $b$ by maximizing the following log-likelihood:
\begin{equation}
\begin{aligned}
\underset{\mathbf{M},b}{\mathop{\max }}\,\left\{ l(\mathbf{M},b)=\sum\limits_{l}{{{h}_{l}}\ln {{p}_{l}}+(1-{{h}_{l}})\ln (1-{{p}_{l}})} \right\}.
\end{aligned}
\end{equation}
Note that $\mathbf{M}$ is not constrained to be positive-definite in LDML.
\par
With the same doublet set, let $\boldsymbol{\alpha}$ be the solution obtained by the kernel logistic model in (12), and $\mathbf{M}$ be the solution of LDML in (24). It is easy to see that:
\begin{equation}
\begin{aligned}
\mathbf{M}=\sum\limits_{l}{{{\alpha }_{l}}({{\mathbf{x}}_{l(i)}}-{{\mathbf{x}}_{l(j)}}){{({{\mathbf{x}}_{l(i)}}-{{\mathbf{x}}_{l(j)}})}^{T}}}.
\end{aligned}
\end{equation}
Thus, LDML is equivalent to kernel logistic regression under the proposed kernel classification framework.
\section{Metric Learning via SVM}
\par
The kernel classification framework proposed in Section III can not only generalize the existing metric learning models, as shown in Section III-D, but also is able to suggest new metric learning models. Actually, for both ITML and LMNN, the positive semi-definite constraint is imposed on $\mathbf{M}$ to guarantee that the learned distance metric is a Mahalanobis metric, which makes the models unable to be solved using the efficient kernel learning toolbox. In this section, a two-step greedy strategy is adopted for metric learning. We first neglect the positive semi-definite constraint and use the SVM toolbox to learn a preliminary matrix $\mathbf{M}$, and then map $\mathbf{M}$ onto the space of positive semi-definite matrices. As examples, we present two novel metric learning methods, namely doublet-SVM and triplet-SVM, based on the proposed framework. Like in conventional SVM, we adopt the $l_2$-norm to regularize $\mathbf{M}$ and employ the hinge loss penalty, and hence the doublet-SVM and triplet-SVM can be efficiently solved by using the standard SVM toolbox.
\subsection{Doublet-SVM}
\par
In doublet-SVM, we set the $l_2$-norm regularizer as ${{r}_{\text{SVM}}}\left( \mathbf{M} \right)=\tfrac{1}{2}\left\| \mathbf{M} \right\|_{F}^{2}$, and set ${{\rho }_{\text{SVM}}}\left( \boldsymbol{\xi } \right)=C\sum\nolimits_{l}{{{\xi }_{l}}}$ as the margin loss term, resulting in the following model:
\begin{equation}
\begin{aligned}
  & \underset{\mathbf{M},b,\boldsymbol{\xi }}{\mathop{\min }}\,\text{ }\frac{1}{2}\left\| \mathbf{M} \right\|_{F}^{2}+C\sum\limits_{l}{{{\xi }_{l}}} \\
 & \text{s}\text{.t}\text{.}\quad {{h}_{l}}\left( {{({{\mathbf{x}}_{l,1}}-{{\mathbf{x}}_{l,2}})}^{T}}\mathbf{M}({{\mathbf{x}}_{l,1}}-{{\mathbf{x}}_{l,2}})+b \right)\ge 1-{{\xi }_{l}} \\
 & \quad \quad {{\xi }_{l}}\ge 0,\text{  }\forall l \\
\end{aligned}
\end{equation}
where ${{\left\| \cdot  \right\|}_{F}}$ denotes the Frobenius norm. The Lagrange dual problem of the above doublet-SVM model is:
\begin{equation}
\begin{aligned}
  & \underset{\boldsymbol{\alpha }}{\mathop{\max }}\,\text{ }-\frac{1}{2}\sum\limits_{i,j}{{{\alpha }_{i}}{{\alpha }_{j}}{{h}_{i}}{{h}_{j}}{{K}_{p}}\left( {{\mathbf{z}}_{i}},{{\mathbf{z}}_{j}} \right)}+\sum\limits_{i}{{{\alpha }_{i}}} \\
 & \text{s}\text{.t}\text{.}\quad 0\le {{\alpha }_{l}}\le C \\
 & \quad \quad \sum\limits_{l}{{{\alpha }_{l}}{{h}_{l}}}=0,\quad \forall l \\
\end{aligned}
\end{equation}
which can be easily solved by many existing SVM solvers such as LibSVM [31]. The detailed deduction of the dual of doublet-SVM can be found in Appendix A.
\subsection{Triplet-SVM}
\par
In triplet-SVM, we also choose ${{r}_{\text{SVM}}}\left( \mathbf{M} \right)=\tfrac{1}{2}\left\| \mathbf{M} \right\|_{F}^{2}$ as the regularization term, and choose ${{\rho }_{\text{SVM}}}\left( \boldsymbol{\xi } \right)=C\sum\nolimits_{l}{{{\xi }_{l}}}$ as the margin loss term. Since the triplets do not have label information, we choose the linear inequality constraints which are adopted in LMNN, resulting in the following triplet-SVM model,
\begin{equation}
\begin{aligned}
   \underset{\mathbf{M},\boldsymbol{\xi }}{\mathop{\min }}\,\text{ } & \frac{1}{2}\left\| \mathbf{M} \right\|_{F}^{2}+C\sum\limits_{l}{{{\xi }_{l}}} \\
  \text{s}\text{.t}\text{.}\quad &
 \begin{array}{c}
 {{({{\mathbf{x}}_{l,1}}-{{\mathbf{x}}_{l,3}})}^{T}}\mathbf{M}({{\mathbf{x}}_{l,1}}-{{\mathbf{x}}_{l,3}})\\
 -{{({{\mathbf{x}}_{l,1}}-{{\mathbf{x}}_{l,2}})}^{T}}\mathbf{M}({{\mathbf{x}}_{l,1}}-{{\mathbf{x}}_{l,2}})
 \end{array}
 \ge 1-{{\xi }_{l}}\text{ } \\
 & {{\xi }_{l}}\ge 0,\text{  }\forall l \\
\end{aligned}.
\end{equation}
\par
Actually, the proposed triplet-SVM can be regarded as a one-class SVM model, and the formulation of triplet-SVM is similar to the one-class SVM in [45]. The dual problem of triplet-SVM is:
\begin{equation}
\begin{aligned}
  & \underset{\boldsymbol{\alpha }}{\mathop{\max }}\,\text{ }-\frac{1}{2}\sum\limits_{i,j}{{{\alpha }_{i}}{{\alpha }_{j}}{{K}_{p}}\left( {{\mathbf{t}}_{i}},{{\mathbf{t}}_{j}} \right)}+\sum\limits_{i}{{{\alpha }_{i}}} \\
 & \text{s}\text{.t}\text{.}\quad 0\le {{\alpha }_{l}}\le C,\quad \forall l \\
\end{aligned}
\end{equation}
which can also be efficiently solved by existing SVM solvers [31]. The detailed deduction of the dual of triplet-SVM can be found in Appendix B.
\subsection{Discussions}
\par
The matrix $\mathbf{M}$ learned by doublet-SVM and triplet-SVM may not be semi-positive definite. To learn a Mahalanobis distance metric, which enforces $\mathbf{M}$ to be semi-positive definite, we can compute the singular value decomposition of $\mathbf{M}=\mathbf{U\Lambda V}$, where $\mathbf{\Lambda }$ is the diagonal matrix of eigenvalues, and then preserve only the positive eigenvalues in $\mathbf{\Lambda }$ to form another diagonal matrix ${{\mathbf{\Lambda }}_{+}}$. Finally, we let ${{\mathbf{M}}_{+}}=\mathbf{U}{{\mathbf{\Lambda }}_{+}}\mathbf{V}$ be the Mahalanobis metric matrix.
\par
The proposed doublet-SVM and triplet-SVM are easy to implement since the use of  $l_2$-norm regularizer and hinge loss penalty allows us to readily employ the available SVM toolbox to solve them. A number of efficient algorithms, e.g., sequential minimal optimization [50], have been proposed for SVM training, making doublet-SVM and triplet-SVM very efficient to optimize. Moreover, using the large-scale SVM training algorithms [46], [47], [51], [53], we can easily extend doublet-SVM and triplet-SVM to deal with large-scale metric learning problems.
\par
A number of kernel methods have been proposed for supervised learning [43]. With the proposed framework, we can easily couple them with the degree-2 polynomial kernel to develop new metric learning approaches for various applications. Similarly, the kernel methods for semi-supervised learning [54], multiple instance learning [55], [56] and multi-task learning [57] can also be adopted for metric learning with the proposed framework.
\section{Experimental Results}
\par
In the experiments, we evaluate the proposed doublet-SVM and triplet-SVM for $k$-NN classification by using the UCI datasets and the handwritten digit datasets. We compare the proposed methods with five representative and state-of-the-art metric learning models, i.e., LMNN [8], ITML [10], LDML [3], neighbourhood component analysis (NCA) [9] and maximally collapsing metric learning (MCML) [2], in terms of classification error rate and training time (in seconds). We implemented doublet-SVM and triplet-SVM based on the popular SVM toolbox LibSVM\footnote{\url{http://www.csie.ntu.edu.tw/~cjlin/libsvm/}} . The source codes of LMNN\footnote{\url{http://www.cse.wustl.edu/~kilian/code/code.html}}, ITML\footnote{\url{http://www.cs.utexas.edu/~pjain/itml/}}, LDML\footnote{\url{http://lear.inrialpes.fr/people/guillaumin/code.php}}, NCA\footnote{\url{http://www.cs.berkeley.edu/~fowlkes/software/nca/}} and MCML\footnote{\url{http://homepage.tudelft.nl/19j49/Matlab_Toolbox_for_Dimensionality_Reduction.html}} are also online available, and we tuned their parameters to get the best results.
\subsection{UCI Dataset Classification}
\par
Ten datasets selected from the UCI machine learning repository [58] are used in the experiment. For the Statlog Satellite, SPECTF Heart and Letter datasets, we use the defined training and test sets to perform the experiment. For the other 7 datasets, we use 10-fold cross validation to evaluate the competing metric learning methods, and the reported error rate and training time are obtained by averaging over the 10 runs. Table I summarizes the basic information of the 10 UCI datasets.
\begin{table*}[!t]

\renewcommand{\arraystretch}{1.3}

\caption{The UCI datasets used in the experiment}
\label{table1}
\begin{center}
%\centering
\begin{tabular}{c|c|c|c|c}
\hline
\bfseries Dataset & \bfseries \# of training samples & \bfseries \# of test samples & \bfseries Feature dimension & \bfseries \# of classes\\
\hline\hline
Parkinsons & 176 & 19 & 22 & 2\\
Sonar & 188 & 20 & 60 & 2\\
Statlog Segmentation & 2 079 & 231 & 19 & 7\\
Breast Tissue & 96 & 10 & 9 & 6\\
ILPD & 525 & 58 & 10 & 2\\
Statlog Satellite & 4 435 & 2 000 & 36 & 6\\
Blood Transfusion & 674 & 74 & 4 & 2\\
SPECTF Heart & 80 & 187 & 44 & 2\\
Cardiotocography & 1 914 & 212 & 21 & 10\\
Letter & 16 000 & 4 000 & 16 & 26\\
\hline
\end{tabular}
\end{center}
\end{table*}
\par
Both doublet-SVM and triplet-SVM involve three hyper-parameters, i.e., $m_1$, $m_2$, and $C$. Using the Statlog Segmentation dataset as an example, we analyze the sensitivity of classification error rate to those hyper-parameters. By setting $m_1 = 1$ and $C = 1$, we investigate the influence of $m_2$ on classification performance. Fig. 1 shows the curves of classification error rate versus $m_2$ for doublet-SVM and triplet-SVM. One can see that both doublet-SVM and triplet-SVM achieve the lowest error rates when $m_2 = 2$. Moreover, the error rates tend to be a little higher when $m_2 > 3$. Thus, we set $m_2$ to 1 $\thicksim$ 3 in our experiments.
\begin{figure}[htb!]
%\vskip 0.2in
\begin{center}
\subfigure[]{
\includegraphics[width=0.45\columnwidth]{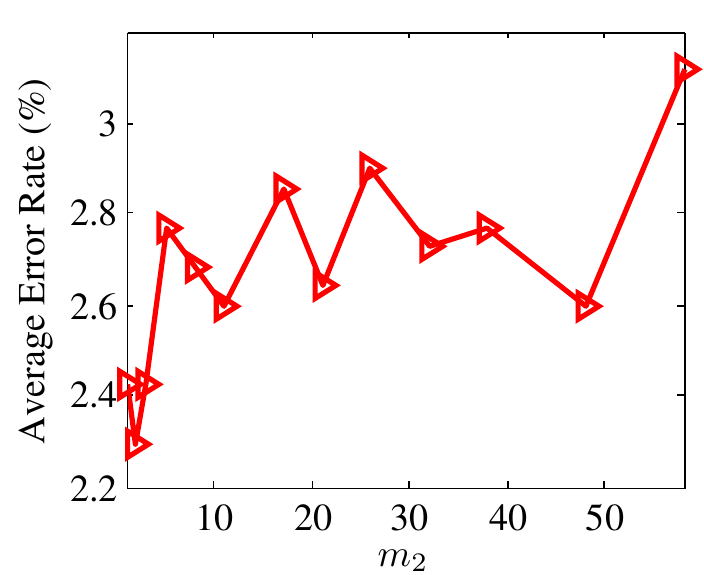}
}
\subfigure[]{
\includegraphics[width=0.45\columnwidth]{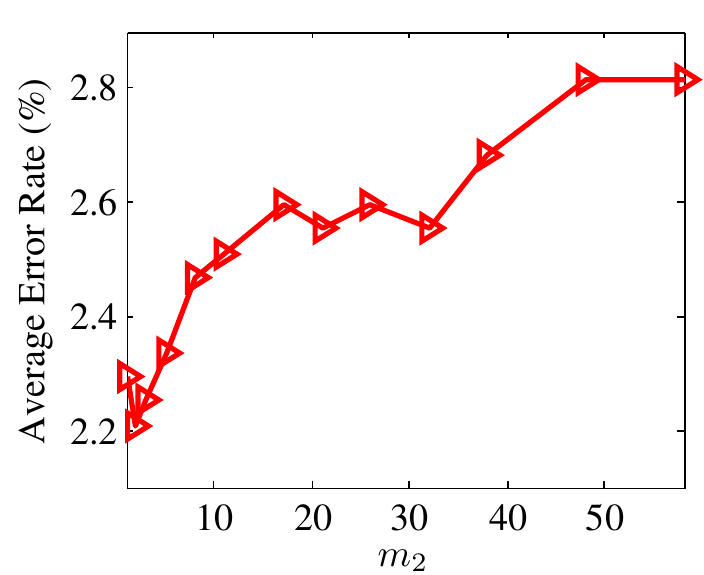}
}
\caption{Classification error rate (\%) versus $m_2$ for (a) doublet-SVM and (b) triplet-SVM with $m_1 = 1$ and $C = 1$.}
\label{fig1}
\end{center}
%\vskip -0.2in
\end{figure}
\par
By setting $m_1 = m_2$, we study the influence of $m_1$ on classification error rate. The curves of error rate versus $m_1\left(=m_2\right)$ for doublet-SVM and triplet-SVM are shown in Fig. 2. One can see that, the lowest classification error is obtained when $m_1 = m_2 = 2$. Thus, we also set $m_1$ to 1 $\thicksim$ 3 in our experiments.
\begin{figure}[htb!]
%\vskip 0.2in
\begin{center}
\subfigure[]{
\includegraphics[width=0.45\columnwidth]{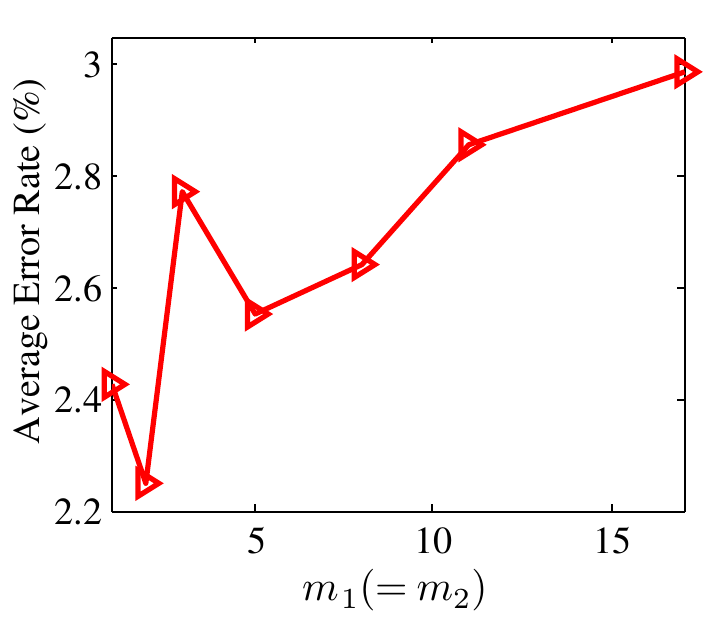}
}
\subfigure[]{
\includegraphics[width=0.45\columnwidth]{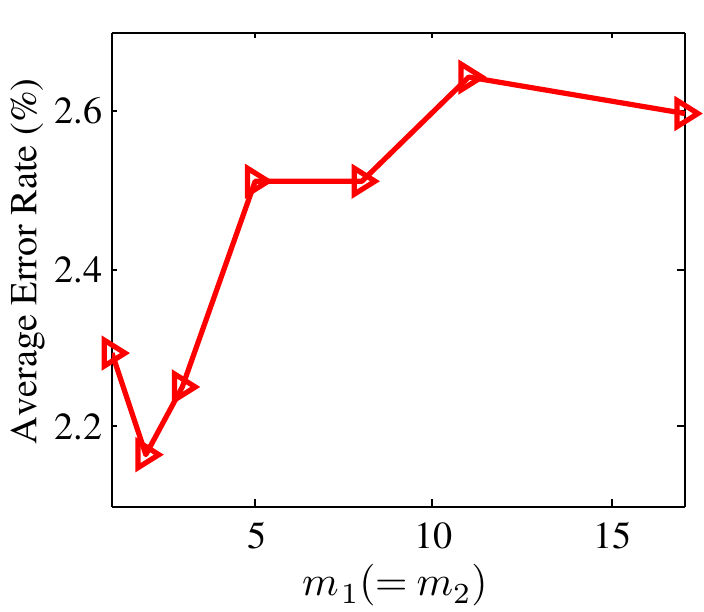}
}
\caption{Classification error rate (\%) versus $m_1(=m_2)$ for (a) doublet-SVM and (b) triplet-SVM with $C = 1$.}
\label{fig2}
\end{center}
%\vskip -0.2in
\end{figure}
\par
We further investigate the influence of $C$ on the classification error rate by fixing $m_1=m_2=2$. Fig. 3 shows the curves of classification error rate versus $C$ for doublet-SVM and triplet-SVM. One can see that the error rate is insensitive to $C$ in a wide range, but it jumps when $C$ is no less than $10^4$ for doublet-SVM and no less than $10^1$ for triplet-SVM. Thus, we set $C<10^4$ for doublet-SVM and $C<10^1$ for triplet-SVM in our experiments.

\begin{figure}[htb!]
%\vskip 0.2in
\begin{center}
\subfigure[]{
\includegraphics[width=0.46\columnwidth]{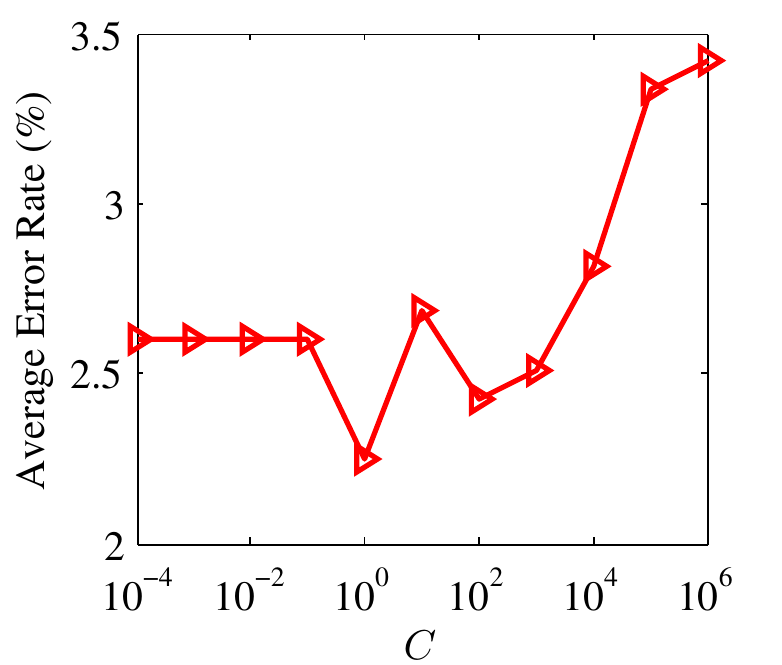}
}
\subfigure[]{
\includegraphics[width=0.46\columnwidth]{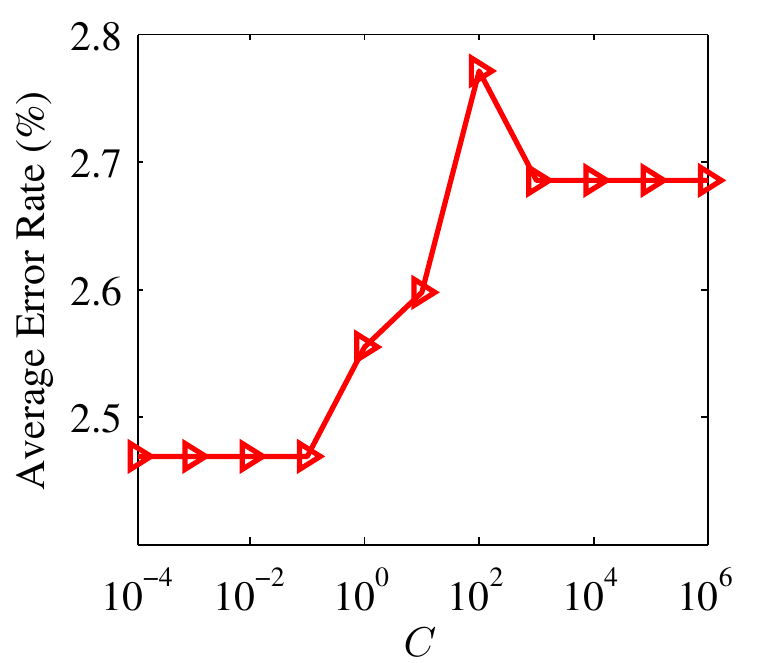}
}
\caption{Classification error rate (\%) versus $C$ for (a) doublet-SVM and (b) triplet-SVM with $m_1=m_2=2$.}
\label{fig3}
\end{center}
%\vskip -0.2in
\end{figure}
\par
Table II lists the classification error rates of the seven metric learning models on the 10 UCI datasets. On the Letter, ILPD and SPECTF Heart datasets, doublet-SVM obtains the lowest error rates. On the Statlog Segmentation dataset, triplet-SVM achieves the lowest error rate. In order to compare the recognition performances of these metric learning models, we list the average ranks of these models in the last row of Table II. On each dataset, we rank the methods based on their error rates, i.e., we assign rank 1 to the best method and rank 2 to the second best method, and so on. The average rank is defined as the mean rank of one method over the 10 datasets, which can provide a fair comparison of the algorithms [59].\\
\begin{table*}[!t]
\renewcommand{\arraystretch}{1.3}
\caption{The classification error rates (\%) and average ranks of the competing methods on the UCI datasets}
\label{table2}
\begin{center}
%\centering
\begin{tabular}{c|c|c|c|c|c|c|c}
\hline
\bfseries Method & \bfseries Doublet-SVM & \bfseries Triplet-SVM & \bfseries NCA & \bfseries LMNN & \bfseries ITML & \bfseries MCML & \bfseries LDML\\
\hline\hline
Parkinsons & 5.68 & 7.89 & \textbf{4.21} & 5.26 & 6.32 & 12.94 & 7.15\\
Sonar & 13.07 & 14.29 & 14.43 & \textbf{11.57} & 14.86 & 24.29 & 22.86\\
Statlog Segmentation & 2.42 & \textbf{2.29} & 2.68 & 2.64 & \textbf{2.29} & 2.77 & 2.86\\
Breast Tissue & 38.37 & 33.37 & \textbf{30.75} & 34.37 & 36.75 & \textbf{30.75} & 48.00\\
ILPD & \textbf{32.09} & 35.16 & 34.79 & 34.12 & 33.59 & 34.79 & 35.84\\
Statlog Satellite & 10.80 & 10.75 & 10.95 & \textbf{10.05} & 11.30 & 15.65 & 15.90\\
Blood Transfusion & 29.47 & 34.37 & \textbf{28.38} & 28.78 & 31.51 & 31.89 & 31.40\\
SPECTF Heart & \textbf{27.27} & 33.69 & 38.50 & 34.76 & 35.29 & 29.95 & 33.16\\
Cardiotocography & 20.71 & 19.34 & 21.84 & \textbf{19.21} & 19.90 & 20.76 & 22.26\\
Letter & \textbf{2.47} & 2.77 & \textbf{2.47} & 3.45 & 2.78 & 4.20 & 11.05\\
\hline\hline
\emph{Average Rank} & \emph{\textbf{2.70}} & \emph{3.70} & \emph{3.40} & \emph{2.80} & \emph{4.00} & \emph{5.00} & \emph{6.00}\\
\hline
\end{tabular}
\end{center}
\end{table*}
\par
From Table II, we can see that doublet-SVM achieves the best average rank and triplet-SVM achieves the fourth best average rank. The results validate that, by incorporating the degree-2 polynomial kernel into the standard (one-class) kernel SVM classifier, the proposed kernel classification based metric learning framework can lead to very competitive classification accuracy with state-of-the-art metric learning methods. It is interesting to see that, although doublet-SVM outperforms triplet-SVM on most datasets, triplet-SVM works better than doublet-SVM on the large datasets like Statlog Segmentation, Statlog Satellite and Cardiotocography, and achieves very close error rate to doublet-SVM on the large dataset Letter. These results may indicate that doublet-SVM is more effective for small scale datasets, while triplet-SVM is more effective for large scale datasets, where each class has many training samples. Our experimental results on the three large scale handwritten digit datasets in Section V-B will further verify this.
\par
Let's then compare the training time of the proposed methods and the competing methods. All the experiments are executed in a PC with 4 Intel Core i5-2410 CPUs (2.30 GHz) and 16 GB RAM. Note that in the training stage, doublet-SVM, ITML, LDML, MCML, and NCA are operated on the doublet set, while triplet-SVM and LMNN are operated on the triplet set. Thus, we compare the five doublet-based metric learning methods and the two triplet-based methods, respectively. Fig. 4 compares the training time of doublet-SVM, ITML, LDML, MCML, and NCA. Clearly, doublet-SVM is always the fastest algorithm and it is much faster than the other four methods. In average, it is 2000 times faster than the second fastest algorithm, ITML. Fig. 5 compares the training time of triplet-SVM and LMNN. One can see that triplet-SVM is about 100 times faster than LMNN on the ten data sets.

\begin{figure}[htb!]
%\vskip 0.2in
\begin{center}
\centerline{\includegraphics[width=\columnwidth]{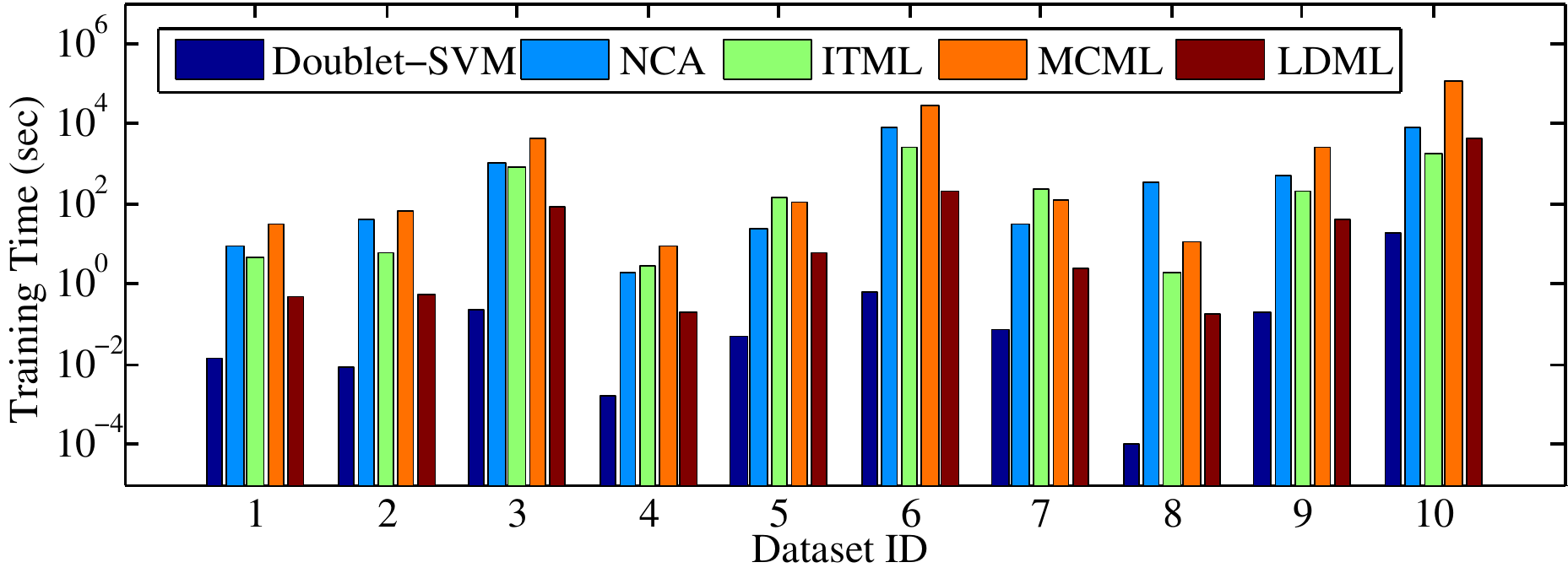}}
\caption{Training time (sec.) of doublet-SVM, NCA, ITML, MCML and LDML. From 1 to 10, the Dataset ID represents Parkinsons, Sonar, Statlog Segmentation, Breast Tissue, ILPD, Statlog satellite, Blood Transfusion, SPECTF Heart, Cardiotocography, and Letter.}
\label{fig4}
\end{center}
%\vskip -0.2in
\end{figure}

\begin{figure}[htb!]
%\vskip 0.2in
\begin{center}
\centerline{\includegraphics[width=\columnwidth]{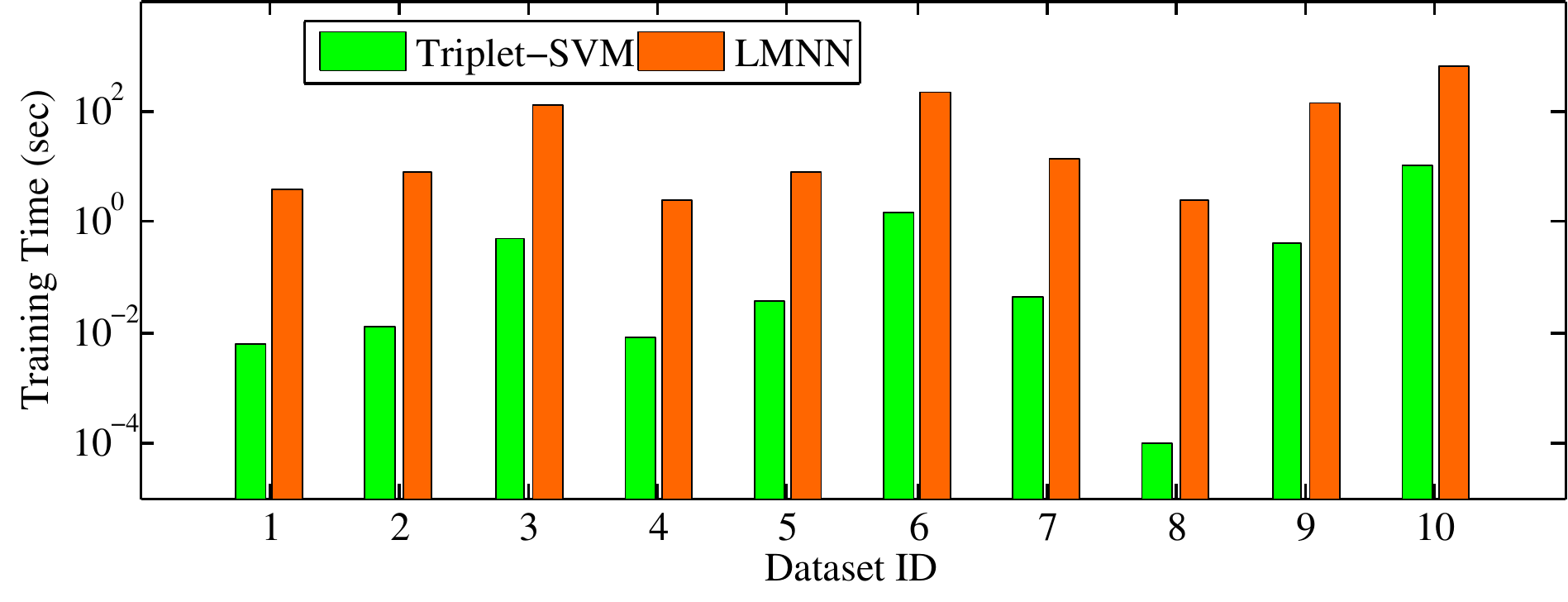}}
\caption{Training time (sec.) of triplet-SVM and LMNN. From 1 to 10, the Dataset ID represents Parkinsons, Sonar, Statlog Segmentation, Breast Tissue, ILPD, Statlog satellite, Blood Transfusion, SPECTF Heart, Cardiotocography, and Letter.}
\label{fig5}
\end{center}
%\vskip -0.2in
\end{figure}

\subsection{Handwritten Digit Recognition}
\par
Apart from the UCI datasets, we also perform experiments on three widely used large scale handwritten digit sets, i.e., MNIST, USPS, and Semeion, to evaluate the performances of doublet-SVM and triplet-SVM. On the MNIST and USPS datasets, we use the defined training and test sets to train the models and calculate the classification error rates. On the Semeion datasets, we use 10-fold cross validation to evaluate the metric learning methods, and the error rate and training time are obtained by averaging over the 10 runs. Table III summarizes the basic information of the three handwritten digit datasets.
\begin{table*}[!t]
\renewcommand{\arraystretch}{1.3}
\caption{The handwritten digits datasets used in the experiments}
\label{table3}
\begin{center}
%\centering
\begin{tabular}{c|c|c|c|c|c}
\hline
\bfseries Dataset & \bfseries \# of training samples & \bfseries \# of test samples & \bfseries Feature dimension & \bfseries PCA dimension & \bfseries \# of classes\\
\hline\hline
MNIST & 60 000 & 10 000 & 784 & 100 & 10\\
USPS & 7 291 & 2 007 & 256 & 100 & 10\\
Semeion & 1 434 & 159 & 256 & 100 & 10\\
\hline
\end{tabular}
\end{center}
\end{table*}

\begin{table*}[!t]
\renewcommand{\arraystretch}{1.3}
\caption{The classification error rates (\%) and average ranks of the competing methods on the handwritten digit datasets}
\label{table4}
\begin{center}
%\centering
\begin{tabular}{c|c|c|c|c|c|c|c}
\hline
\bfseries Dataset & \bfseries Doublet-SVM & \bfseries Triplet-SVM & \bfseries NCA & \bfseries LMNN & \bfseries ITML & \bfseries MCML & \bfseries LDML\\
\hline\hline
MNIST & 3.19 & 2.92 & 5.46 & \textbf{2.28} & 2.89 & - & 6.05\\
USPS & \textbf{5.03} & 5.23 & 5.68 & 5.38 & 6.63 & 5.08 & 8.77\\
Semeion & 5.09 & \textbf{4.71} & 8.60 & 6.09 & 5.71 & 11.23 & 11.98\\
\hline\hline
\emph{Average Rank} & \emph{2.33} & \emph{\textbf{2.00}} & \emph{4.67} & \emph{2.67} & \emph{3.33} & - & \emph{6.00}\\
\hline
\end{tabular}
\end{center}
\end{table*}
\par
As the dimensions of digit images are relatively high, PCA is utilized to reduce the feature dimension. The metric learning models are trained in the PCA subspace. Table IV lists the classification error rates on the handwritten digit datasets. On the MNIST dataset, LMNN achieves the lowest error rate; on the USPS dataset, doublet-SVM achieves the lowest error rate; and on the Semeion dataset, triplet-SVM obtains the lowest error rate. We do not report the error rate of MCML on the MNIST dataset because MCML requires too large memory space (more than 30 GB) on this dataset and cannot be run in our PC.

\par
The last row of Table IV lists the average ranks of the seven metric learning models. We can see that triplet-SVM can achieve the best average rank, and doublet-SVM achieves the second best average rank. The results further validate that on large scale datasets where each class has sufficient number of training samples, triplet-SVM would be superior to doublet-SVM and the competing methods.
\par
We then compare the training time of these metric learning methods. All the experiments are executed in the same PC as the experiments in Section V-A. We compare the five doublet-based metric learning methods and the two triplet-based methods, respectively. Fig. 6 shows the training time of doublet-SVM, ITML, LDML, MCML, and NCA. We can see that doublet-SVM is much faster than the other four methods. In average it is 2000 times faster than the second fastest algorithm, ITML. Fig. 7 shows the training time of triplet-SVM and LMNN. One can see that triplet-SVM is about 100 times faster than LMNN on the three datasets.

\begin{figure}[htb!]
%\vskip 0.2in
\begin{center}
\centerline{\includegraphics[width=0.9\columnwidth]{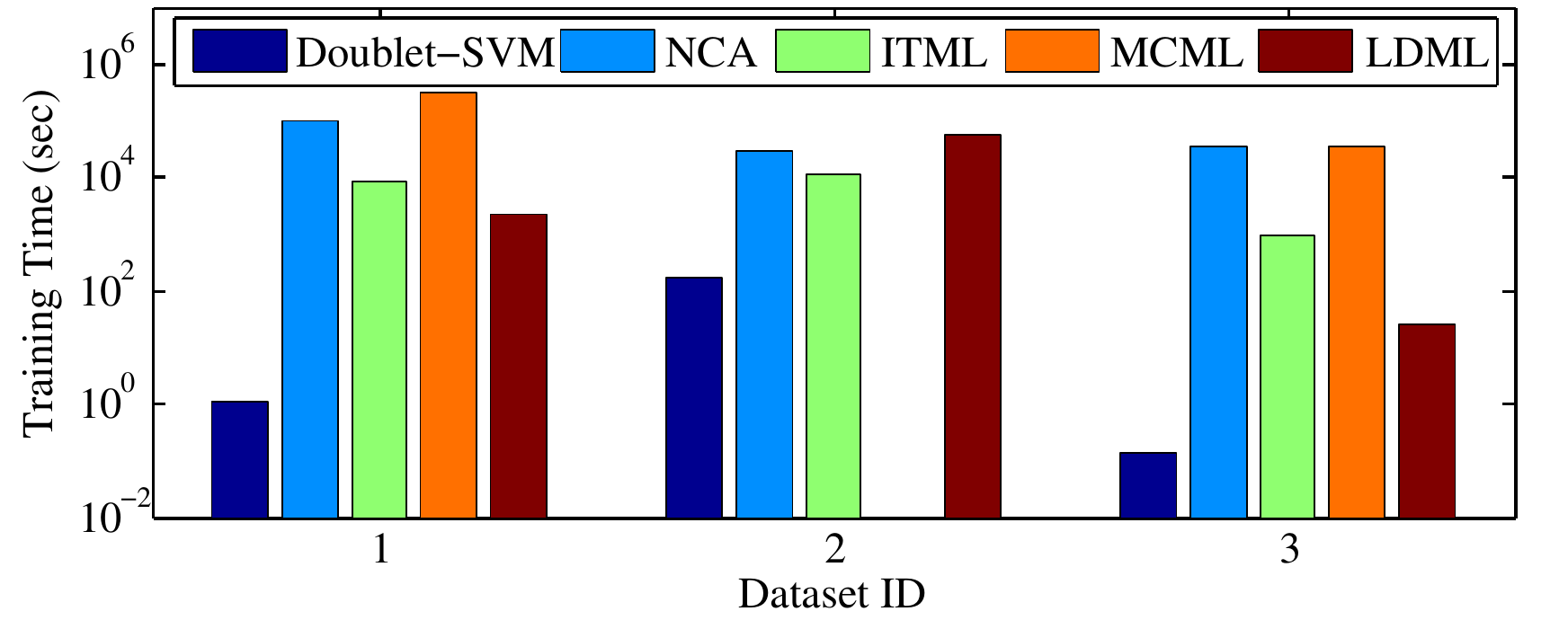}}
\caption{Training time (sec.) of doublet-SVM, NCA, ITML, MCML and LDML. From 1 to 3, the Dataset ID represents USPS, MNIST and Semeion.}
\label{fig4}
\end{center}
%\vskip -0.2in
\end{figure}

\begin{figure}[htb!]
%\vskip 0.2in
\begin{center}
\centerline{\includegraphics[width=0.5\columnwidth]{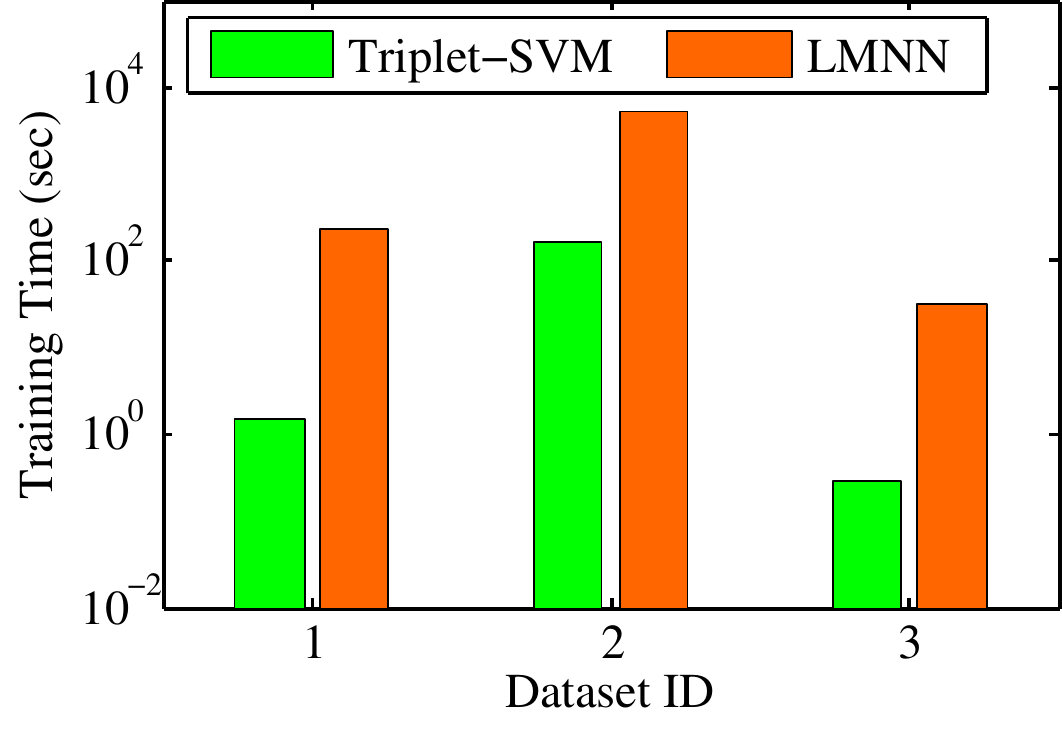}}
\caption{Training time (sec.) of triplet-SVM and LMNN. From 1 to 3, the Dataset ID represents USPS, MNIST and Semeion.}
\label{fig5}
\end{center}
%\vskip -0.2in
\end{figure}

\section{Conclusion}
\par
In this paper, we proposed a general kernel classification framework for distance metric learning. By coupling a degree-2 polynomial kernel with some kernel methods, the proposed framework can unify many representative and state-of-the-art metric learning approaches such as LMNN, ITML and LDML. The proposed framework also provides a good platform for developing new metric learning algorithms. As examples, two metric learning methods, i.e., doublet-SVM and triplet-SVM, were developed and they can be efficiently solved by the standard SVM solvers. Our experimental results on the UCI datasets and handwritten digit datasets showed that doublet-SVM and triplet-SVM are much faster than state-of-the-art methods in terms of training time, while they achieve very competitive results in terms of classification error rate.

% or


\appendices
\section{The dual of doublet-SVM}
\par
According to the original problem of doublet-SVM in (26), its Lagrangian can be defined as follows:
\begin{equation}
\begin{aligned}
& L\left( \mathbf{M},b,\boldsymbol{\xi },\boldsymbol{\alpha },\boldsymbol{\beta } \right)=\frac{1}{2}\left\| \mathbf{M} \right\|_{F}^{2}+C\sum\limits_{l}{{{\xi }_{l}}}\\
& -\sum\limits_{l}{{{\alpha }_{l}}\left[
\begin{array}{c}
 {{h}_{l}}\left( {{({{\mathbf{x}}_{l,1}}-{{\mathbf{x}}_{l,2}})}^{T}}\mathbf{M}({{\mathbf{x}}_{l,1}}-{{\mathbf{x}}_{l,2}})+b \right)-1+{{\xi }_{l}}
 \end{array}
 \right]}\\
& -\sum\limits_{l}{{{\beta }_{l}}{{\xi }_{l}}}
\end{aligned}
\end{equation}
where $\boldsymbol{\alpha }$ and $\boldsymbol{\beta }$ are the Lagrange multipliers which satisfy ${{\alpha }_{l}}\ge 0$ and ${{\beta }_{l}}\ge 0,\ \forall l$. To convert the original problem to its dual, we let the derivative of the Lagrangian with respect to $\mathbf{M}$, $b$ and $\boldsymbol{\xi }$ to be $\mathbf{0}$:
\begin{equation}
\begin{aligned}
&\frac{\partial L\left( \mathbf{M},b,\boldsymbol{\xi },\boldsymbol{\alpha },\boldsymbol{\beta } \right)}{\partial \mathbf{M}}=\mathbf{0}\Rightarrow\\
& \mathbf{M}-\sum\limits_{l}{{{\alpha }_{l}}{{h}_{l}}\left( {{\mathbf{x}}_{l,1}}-{{\mathbf{x}}_{l,2}} \right){{\left( {{\mathbf{x}}_{l,1}}-{{\mathbf{x}}_{l,2}} \right)}^{T}}}=\mathbf{0}
\end{aligned}
\end{equation}

\begin{equation}
\begin{aligned}
\frac{\partial L\left( \mathbf{M},b,\boldsymbol{\xi },\boldsymbol{\alpha },\boldsymbol{\beta } \right)}{\partial b}=0\Rightarrow \sum\limits_{l}{{{\alpha }_{l}}{{h}_{l}}}=0
\end{aligned}
\end{equation}

\begin{equation}
\begin{aligned}
&\frac{\partial L\left( \mathbf{M},b,\boldsymbol{\xi },\boldsymbol{\alpha },\boldsymbol{\beta } \right)}{\partial {{\xi }_{l}}}=0\Rightarrow\\
& C-{{\alpha }_{l}}-{{\beta }_{l}}=0\Rightarrow 0<{{\alpha }_{l}}<C,\ \forall l
\end{aligned}
\end{equation}
\par
Equation (31) implies the relationship between $\mathbf{M}$ and $\boldsymbol{\alpha }$ as follows:
\begin{equation}
\begin{aligned}
\mathbf{M}=\sum\limits_{l}{{{\alpha }_{l}}{{h}_{l}}\left( {{\mathbf{x}}_{l,1}}-{{\mathbf{x}}_{l,2}} \right){{\left( {{\mathbf{x}}_{l,1}}-{{\mathbf{x}}_{l,2}} \right)}^{T}}}
\end{aligned}
\end{equation}
\par
Substituting (31)$\thicksim$(33) back into the Lagrangian, we get the Lagrange dual problem of doublet-SVM as follows:
\begin{equation}
\begin{aligned}
  & \underset{\boldsymbol{\alpha }}{\mathop{\max }}\,\text{ }-\frac{1}{2}\sum\limits_{i,j}{{{\alpha }_{i}}{{\alpha }_{j}}{{h}_{i}}{{h}_{j}}{{K}_{p}}\left( {{\mathbf{z}}_{i}},{{\mathbf{z}}_{j}} \right)}+\sum\limits_{i}{{{\alpha }_{i}}} \\
 & \text{s}\text{.t}\text{.}\quad 0\le {{\alpha }_{l}}\le C \\
 & \quad \quad \sum\limits_{l}{{{\alpha }_{l}}{{h}_{l}}}=0,\quad \forall l \\
\end{aligned}
\end{equation}

\section{The dual of triplet-SVM}
\par
According to the original problem of triplet-SVM in (28), its Lagrangian can be defined as follows:
\begin{equation}
\begin{aligned}
  & L\left( \mathbf{M},\boldsymbol{\xi },\boldsymbol{\alpha },\boldsymbol{\beta } \right)=\frac{1}{2}\left\| \mathbf{M} \right\|_{F}^{2}+C\sum\limits_{l}{{{\xi }_{l}}}\\
  & -\sum\limits_{l}{{{\alpha }_{l}}\left[
  \begin{array}{c}
  {{({{\mathbf{x}}_{l,1}}-{{\mathbf{x}}_{l,3}})}^{T}}\mathbf{M}({{\mathbf{x}}_{l,1}}-{{\mathbf{x}}_{l,3}})\\
  -{{({{\mathbf{x}}_{l,1}}-{{\mathbf{x}}_{l,2}})}^{T}}\mathbf{M}({{\mathbf{x}}_{l,1}}-{{\mathbf{x}}_{l,2}})
  \end{array}
  \right]} \\
 & +\sum\limits_{l}{{{\alpha }_{l}}}-\sum\limits_{l}{{{\alpha }_{l}}{{\xi }_{l}}}-\sum\limits_{l}{{{\beta }_{l}}{{\xi }_{l}}}
\end{aligned}
\end{equation}
where $\boldsymbol{\alpha }$ and $\boldsymbol{\beta }$ are the Lagrange multipliers, which satisfy ${{\alpha }_{l}}\ge 0$ and ${{\beta }_{l}}\ge 0,\ \forall l$. To convert the original problem to its dual, we let the derivative of the Lagrangian with respect to $\mathbf{M}$ and $\boldsymbol{\xi }$ to be $\mathbf{0}$:
\begin{equation}
\begin{aligned}
& \frac{\partial L\left( \mathbf{M},\boldsymbol{\xi },\boldsymbol{\alpha },\boldsymbol{\beta } \right)}{\partial \mathbf{M}}=\mathbf{0}\Rightarrow\\
& \mathbf{M}-\sum\limits_{l}{{{\alpha }_{l}}\left[
\begin{array}{c}
 \left( {{\mathbf{x}}_{l,1}}-{{\mathbf{x}}_{l,3}} \right){{\left( {{\mathbf{x}}_{l,1}}-{{\mathbf{x}}_{l,3}} \right)}^{T}}\\
 -\left( {{\mathbf{x}}_{l,1}}-{{\mathbf{x}}_{l,2}} \right){{\left( {{\mathbf{x}}_{l,1}}-{{\mathbf{x}}_{l,2}} \right)}^{T}}
 \end{array}
 \right]}=\mathbf{0}
\end{aligned}
\end{equation}

\begin{equation}
\begin{aligned}
&\frac{\partial L\left( \mathbf{M},\boldsymbol{\xi },\boldsymbol{\alpha },\boldsymbol{\beta } \right)}{\partial {{\xi }_{l}}}=0\Rightarrow \\
& C-{{\alpha }_{l}}-{{\beta }_{l}}=0\Rightarrow 0<{{\alpha }_{l}}<C,\ \forall l
\end{aligned}
\end{equation}
\par
Equation (37) implies the relationship between $\mathbf{M}$ and $\boldsymbol{\alpha }$ as follows:
\begin{equation}
\begin{aligned}
\mathbf{M}=\sum\limits_{l}{{{\alpha }_{l}}\left[
\begin{array}{c}
 \left( {{\mathbf{x}}_{l,1}}-{{\mathbf{x}}_{l,3}} \right){{\left( {{\mathbf{x}}_{l,1}}-{{\mathbf{x}}_{l,3}} \right)}^{T}}\\
 -\left( {{\mathbf{x}}_{l,1}}-{{\mathbf{x}}_{l,2}} \right){{\left( {{\mathbf{x}}_{l,1}}-{{\mathbf{x}}_{l,2}} \right)}^{T}}
 \end{array}
 \right]}
\end{aligned}
\end{equation}
\par
Substituting (37) and (38) back into the Lagrangian, we get the Lagrange dual problem of triplet-SVM as follows:
\begin{equation}
\begin{aligned}
  \underset{\boldsymbol{\alpha }}{\mathop{\max }}\quad & -\frac{1}{2}\sum\limits_{i,j}{{{\alpha }_{i}}{{\alpha }_{j}}{{K}_{p}}\left( {{\mathbf{t}}_{i}},{{\mathbf{t}}_{j}} \right)}+\sum\limits_{i}{{{\alpha }_{i}}} \\
  \text{s}\text{.t}\text{.}\quad & 0\le {{\alpha }_{l}}\le C,\quad \forall l \\
\end{aligned}
\end{equation}

\section*{Acknowledgment}
\par
This work was supported in part by the National Natural Science Foundation of China under Grant 61271093 and Grant 61001037, the Hong Kong Scholar Program, and the Program of Ministry of Education for New Century Excellent Talents.

\ifCLASSOPTIONcaptionsoff
  \newpage
\fi

\end{document}